\documentclass[journal]{IEEEtran}
\usepackage{graphicx} 
\usepackage{hyperref}
\usepackage{subcaption}
\usepackage{amsmath}
\usepackage{multirow}
\usepackage{svg}
\usepackage{fontspec}
\usepackage{seqsplit} 
\usepackage{xspace}

\newfontfamily\consolas[
  Path=consola_font/, 
  UprightFont = CONSOLA.TTF,
  BoldFont    = CONSOLAB.TTF,
  ItalicFont  = CONSOLAZ.TTF
]{Consolas}
\newcommand{\code}[1]{\texttt{\seqsplit{#1}}}
\newcommand{\datasetname}{NFI\_FARED\xspace}

\UseRawInputEncoding

\begin{document}

\title{Forensic Activity Classification Using Digital Traces from iPhones: A Machine Learning-based Approach}

\author{
    \IEEEauthorblockN{
        Conor McCarthy\IEEEauthorrefmark{1}, 
        Jan Peter van Zandwijk\IEEEauthorrefmark{2}\IEEEauthorrefmark{3},
        Marcel Worring\IEEEauthorrefmark{1},
        Zeno Geradts\IEEEauthorrefmark{1}\IEEEauthorrefmark{2}
    }
    \\
    \IEEEauthorblockA{\IEEEauthorrefmark{1}University of Amsterdam, Amsterdam, The Netherlands \\
    Email: \{c.t.mccarthy, m.worring\}@uva.nl}
    \\
    \IEEEauthorblockA{\IEEEauthorrefmark{2}Netherlands Forensic Institute (NFI), The Hague, The Netherlands \\
    Email: \{j.p.van.zandwijk, z.geradts\}@nfi.nl}
    \\
    \IEEEauthorblockA{\IEEEauthorrefmark{3}Amsterdam University of Applied Sciences, Faculty of Technology, Amsterdam, The Netherlands}
}

\maketitle

\begin{abstract}

Smartphones and smartwatches are ever-present in daily life, and provide a rich source of information on their users' behaviour. In particular, digital traces derived from the phone's embedded movement sensors present an opportunity for a forensic investigator to gain insight into a person's physical activities. In this work, we present a machine learning-based approach to translate digital traces into likelihood ratios (LRs) for different types of physical activities. Evaluating on a new dataset, NFI\_FARED, which contains digital traces from four different types of iPhones labelled with 19 activities, it was found that our approach could produce useful LR systems to distinguish 167 out of a possible 171 activity pairings. The same approach was extended to analyse likelihoods for multiple activities (or groups of activities) simultaneously and create activity timelines to aid in both the early and latter stages of forensic investigations. The dataset\footnote{\url{https://huggingface.co/NetherlandsForensicInstitute/datasets}} and all code\footnote{\url{https://github.com/Con-or-McCarthy/Data2Activity_1}} required to replicate the results have also been made public to encourage further research on this topic. 

\end{abstract}

\section*{Introduction} \label{section:introduction}

Smartphones and smartwatches have become a central component of modern life, integrated into almost all parts of daily routines. They are essential not only for communication, but also internet access, navigation, payments, and translation. Smart devices have consequently become indispensable, and people typically carry at least one on their person at all times.
From a digital forensic perspective, the strong integration of digital devices into everyday life offers great investigative opportunities. Given the wide range of capabilities of modern devices, they contain a diverse set of data and could provide unique information that is not obtainable from other types of forensic evidence. We concentrate on the embedded sensors like accelerometers and gyroscopes, the raw data of which is processed into detailed information about physical activities, and stored on the device in the form of digital traces \cite{van_zandwijk_digital_2023}. Significant research exists on the availability and extraction of digital traces from iPhones, whose traces will be the focus of this work.
\par
The potential of such digital traces from iPhones to inform forensic investigators about the user's physical activities has been studied, specifically using information on registrations of steps, distances and floors \cite{van_zandwijk_iphone_2019, van_zandwijk_have_2023, van_zandwijk_phone_2021}, to estimate the user's activities in a certain window of time. While such registrations can be easily interpreted, many other available registrations are less self-explanatory but nonetheless contain additional information that could help to identify physical activities. Machine learning techniques are well suited to this case, as these can take advantage of any patterns present in all available variables in the digital traces to classify activities, and consequently could increase the number of discernible activities. 
\par
For use in forensic investigations, and especially if being used as evidence in court, it is beneficial if the evidential strength of activity classifications made by machine learning algorithms can be quantified. An appropriate approach to achieve this is through the construction of a Likelihood Ratio (LR) system. The benefit of producing an LR in place of a typical classification is that it can be placed in context alongside other evidence to create a more complete picture of events. An LR also is a more transparent measure of evidential strength which aids reliability from a legal perspective. Therefore, our approach produces outputs in the form of LRs.
\par
In this work, we present a machine learning approach for the interpretation of digital traces related to physical activities to estimate the LR of certain activities being performed in a specific time interval. The method uses timestamped digital traces extracted from smart devices and outputs an LR. The model can output LRs for two specific activities (the binary case), stating how much more likely the data was generated by the user performing activity A rather than activity B. For example, ``The traces are fifty times more likely if the user  was driving a car than sitting at home". This form of LR is typical when presenting findings as evidence. Output can also be in the multiclass case, comparing the likelihood of multiple possible activities at once, loosening the necessary assumptions required for producing an LR of only two classes. This also enables the construction of a timeline of most likely activities at each moment, which can be informative to investigators. 
\section{Related Work} \label{section:related_work}



LRs are a useful measure of evidential strength. They present the degree of support for one hypothesis versus another and provide a logically correct method to assist the investigator in assessing their uncertainty \cite{aitken_statistics_2004, taroni_dismissal_2016, evett_logical_2015, evett_towards_1998, robertson_interpreting_2016}. Usage of LR in forensic applications is well developed, with established methods for evaluation \cite{brummer_application-independent_2006, brummer_calibration_2006, banks_handbook_2020} and improving stability \cite{vergeer_numerical_2016}, and has been encouraged by institutions such as ENFSI as a suitable way to report evidence \cite{aitken_enfsi_2015}. As such, forensic likelihood ratios have been utilised in domains such as speaker analysis \cite{xiao_wang_effect_2019, sergidou_frequent-words_2023}, face recognition \cite{macarulla_rodriguez_likelihood_2020}, DNA \cite{collins_likelihood_1994}, drug comparison \cite{bolck_evaluating_2015}, and glass analysis \cite{van_es_implementation_2017}, among others. 
\par
Previous work on producing evidence from digital traces focuses on a single variable such as distance \cite{vink_likelihood_2022} or location \cite{spichiger_likelihood_2023}, and assessing the LR of variable values being generated from different ground truth scenarios. To the best of the authors' knowledge, there is no existing work using digital traces to estimate the likelihood of quantities outside of those reported directly by the variables.
\par
Using body-worn sensor data to classify activities is a large field of research in the machine learning community \cite{singh_deep_2021, mccarthy_hi-oscar_2025, gu_survey_2022, mekruksavanich_resnet-se_2022}, however the data is almost exclusively raw data in the form of high-frequency time series, and techniques do not easily translate to digital trace data.
\section{Collection of Digital Trace Data} \label{section:data}

\paragraph{Participants}
Fourteen test subjects (6 females, 8 males, mean age 26.6y, standard deviation 8.8y) participated in the data collection experiments for this study. Each participant signed an informed consent form, agreeing to the collection and spreading of their data anonymously.  For each subject, data was collected during multiple experimental sessions spread over several days. Due to scheduling conflicts, three participants could not participate in all sessions, and three additional subjects were recruited to complete those sessions in their place. This effectively provides a complete experimental dataset for eleven participants.

\paragraph{Smartphones}
In the experiments, four different iPhones were carried simultaneously by the test subjects. The models and iOS versions of these iPhones are: 6+ (11.4.1), 7 (14.7.1), 11 (13.1.1), and XR (15.4.1). 

\paragraph{Experimental protocol}
In total each subject performed 19 different activities. These activities were carried out in four separate measurement sessions. Session 1 included movement activities, session 2 contained elevation change activities, session 3 forensically relevant dynamic activities, and session 4 consisted of transport activities. The order in which the different activities were performed within each session was randomised for each subject. 
\par
During sessions, subjects wore the iPhones at the following locations:  front trouser pocket, back trouser pocket, breast pocket, backpack, and the hand.  Phone placement was randomised across subjects, but was for each subject kept the same during all four experimental sessions.
\par
Sessions 1, 2 and 3 were performed under controlled conditions. Between subsequent activities, the subject would pause for one minute by either sitting or standing. To encourage diversity in their execution, while carrying out the activities, the subjects were given as few instructions as possible. At the end of each session, the subject completed a ``free-living" section, in which they would perform the activities from that session in any order and for however long they wished without breaks. For each of the activities carried out by the subject, the start and end time of that activity was recorded by the experimenter. Details of each of the sessions is described in more detail below.

\paragraph{Session 1: Movement activities}
The movements walking, running and cycling were  performed multiple times for known distances. For running and walking there are two distances: $\sim240$m and $\sim90$m, each performed twice. For cycling, the subjects cycled a set number of loops around the car park of the experimental location three times. The number of loops each were varied between two, three, and four. During pauses between trials, the subject sat on a bench without backrest. 

\paragraph{Session 2: Elevation Changes}
The movements stair, escalator and elevator were performed four times each going upstairs and downstairs. The stair climbing and escalator movements were performed at different velocities. For stair climbing the subject was instructed to walk or to run on the stairs. For the escalator, the subject was instructed to stand or walk on the escalator. After each movement the participant would either stand or sit for one minute at the end of the stairs/escalator. During pauses between trial,  the subject sat on a bench with backrest. 

\paragraph{Session 3: Forensically relevant dynamic movements }
This session includes the activities kicking, throwing, punching and dragging a heavy object. These activities were performed for time intervals of 10, 20 and 30 seconds. Kicking and punching were performed on a punching bag and was alternated between the dominant and non-dominant arm or leg. For the throwing task, the subject threw a ball weighing 1 kg using their dominant arm down a hallway to the experimenter, who rolled it back to the subject to pick it up and throw again until the task was complete. For the dragging activity the subject dragged the punching bag weighing 34 kg through a hallway. 

\paragraph{Session 4: Transportation }
In this session, the transport modes riding the train, tram, and in a car were carried out by the subjects while travelling home from the experimental location. If the subject had not used one or more of the modes of transport in their trip home, arrangements were made to carry out the missing modes in an additional session. In session 4, the participants recorded the start and end times of each travel activity by themselves.

\paragraph{Data acquisition and processing}
After an experimental session, a full-filesystem extraction of the data of each of the four iPhone was produced using commercial forensic equipment present at Netherlands Forensic Institute. From these extractions, the files \code{\consolas healthdb\_secure.sqlite} and \code{\consolas cache\_encryptedC.db} were exported. These files are known to contain information related to physical activities performed by the phone's user \cite{van_zandwijk_iphone_2019, van_zandwijk_phone_2021, van_zandwijk_have_2023}. 
\par
After export, post-processing of the data contained in the exported files was performed. In this post-processing, data from various Tables from the databases were combined into a single dataset. For the file \code{\consolas cache\_encryptedC.db}, variables were extracted from the tables \code{\consolas StepCountHistory}, \code{\consolas MotionStateHistory} and \code{\consolas NatalieHistory} \cite{van_zandwijk_phone_2021}. For the file \code{\consolas healthdb\_secure.sqlite}, variables were extracted from tables \code{\consolas samples} and \code{\consolas quantity\_samples} \cite{van_zandwijk_iphone_2019}. 
\par
Extracted variables could either be categorical or numerical, and numerical variables could be further separated between cumulative and non-cumulative, where cumulative variables have an always increasing count, and non-cumulative do not. An example of a categorical variable is \code{\consolas type} from the table \code{\consolas MotionStateHistory} in \code{\consolas cache\_encryptedC.db}, which can take one of six different values. An example of a cumulative numerical variable is \code{\consolas floorsAcended} from the table \code{\consolas StepCountHistory} in \code{\consolas cache\_encryptedC.db}, whose value either remains the same or increments. The variable \code{\consolas mets} from the table \code{\consolas NatalieHistory} in \code{\consolas cache\_encryptedC.db}, is an example of a non-cumulative variable, which takes on a new, continuous value at each reading.
\par
After post-processing, data was further aggregated into one minute intervals. In the case there were multiple registrations of a variable being within the same one minute interval, we used the modal variable for categorical variables, the sum of all values for cumulative variables and the mean for non-cumulative variables.  In the case of a variable having no registration in a one minute interval, a missing value (NA) was imputed. NA values are used instead of zero values because a registered zero value has a different interpretation to no registration whatsoever. Ground truth values of activities executed were added to the data before aggregation. During aggregation, if two or more unique activity labels were present in a single minute of data, the minute would be split, with the corresponding variable readings aggregated to the appropriate label. This means that a one minute timestamp could appear more than once in the final dataset, each one containing data for less than a minute of activity. This approach was chosen over simply labelling the minute with the majority class to maximise data availability, particularly for activities that spanned a shorter duration, such as the \textit{dynamic} activities.
\par
Henceforth, we will refer to this dataset as \datasetname. Our dataset is fully publicly available for download and scripts for processing are available on the project's GitHub.
\section{Method} \label{section:method}
\begin{figure*}[ht]
    \centering
    \includegraphics[width=0.75\textwidth]{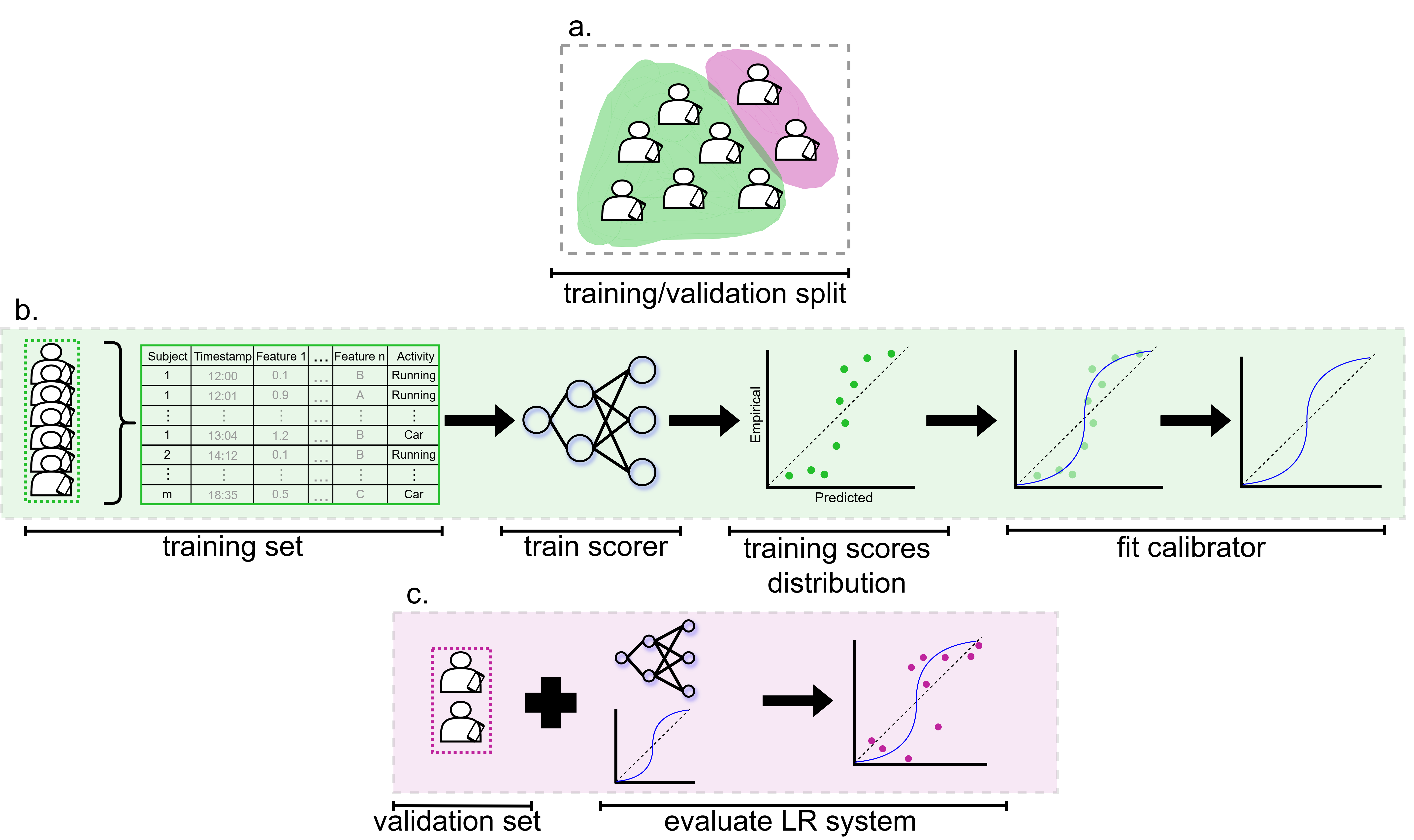}
    \caption{Overview of method employed to train and evaluate an LR system using the proposed approach. a) Subjects are split between training (green) and validation (purple) to prevent data leakage. b) Data from training subjects is collected into the training set, containing the activity classes of interest, which is used to train the scorer (CatBoost\cite{prokhorenkova_catboost_2018}). After training, the scores for the entire training set are calculated to produce the training scores distribution, on which we fit the calibrator. The combination of scorer + calibrator is the LR system c) Validation data is taken from the validation subjects and fed through the LR system from b. The resulting likelihood ratios can then be evaluated using $C_{llr}$, PAV plots, Tippet plots, etc. }
    \label{fig:method_overview}
\end{figure*}

From the data described in \nameref{section:data} above, a machine learning approach consisting of a scorer and calibrator is created which can return an LR for any inputted sample of data (e.g. one minute from \datasetname). The choice of scorer, as well as the procedure followed to create and evaluate the LR system are detailed in the following sections. It should be noted that while results are based on the \datasetname dataset, the described approach can be applied to other datasets containing timestamped traces and also extended in the event of additional traces becoming available to forensic practitioners. 

\subsection{Scorer Description}\label{section:method:model_description}

A score-based LR system works by assigning a single number to each observation to distinguish between $H_1$-true and $H_2$-true samples, where a higher score is more likely to come from a $H_1$-true sample, and a lower score from a $H_2$-true sample, in the binary case. In the multiclass case, $K$ numbers are outputted, one for each of the $K$ distinct hypotheses. Various options are available to achieve this, both in the form of fixed score functions and statistical models. Machine learning approaches learn parameters from the training data rather than relying on expert knowledge to find a good fit to a probability distribution. This increases flexibility in the features which can be used as input, since interpretability of inputs is not a requirement, unlike when crafting specialised scoring rules. Given a sufficiently representative training set, an adequately expressive model, and effective optimization, a machine learning model can learn parameters that approximate the optimal solution with respect to the underlying data distribution.
\par
The digital trace data also exhibits certain characteristics that require consideration when selecting an appropriate model, namely the presence of many missing values, and features being a mix of categorical and numerical. The large amount of missing values arise as a consequence of the the iPhone having irregular and inconsistent sampling rates between variables. Certain variables, such as \code{\consolas floorsAscended}, log a value only when a floor change is detected, resulting in a missing value in all samples without a floor change detected. Data imputation or removal is a standard method of handling missing values, however the missing values can still be informative to the investigator. For example, the lack of a \code{\consolas floorsAscended} registration could increase the likelihood of activity \textit{driving}, while decreasing the likelihood of \textit{upstairs}. Therefore, a model that can natively handle missing values is needed to leverage the maximum information. Similarly, the data containing both numerical and categorical features necessitates a model which can process both data types, to avoid potential bias from data transformations. 
\par
To satisfy these conditions, the tree-based machine learning model CatBoost \cite{prokhorenkova_catboost_2018} was selected. CatBoost uses gradient boosting, in which many decision trees are constructed iteratively and incorporated into the ensemble model. Each tree is fit on the residuals of the model in the last step to improve results. CatBoost natively deals with categorical variables without the need for preprocessing by using ordered encoding, in which target statistics from all the rows prior to a data point are considered to calculate a value to replace the categorical feature. CatBoost is faster than other gradient boosting methods such as XGBoost \cite{chen_xgboost_2016} and gradient boosting approaches are state-of-the-art on tabular data, outperforming much more complex deep learning algorithms \cite{borisov_deep_2022, mcelfresh_when_2023}. 
\subsection{Model Training}\label{section:method:model_training}

As illustrated in Figure \ref{fig:method_overview}a, the dataset is first split into a training and validation sets, with each subject assigned entirely to one set. This prevents data leakage resulting from subjects being shared between the training and validation sets. From the training subjects, all samples of the data relating to the relevant activities are placed into the training set. In Figure \ref{fig:method_overview}b, this is shown for the case of distinguishing activity \textit{car} from activity \textit{running}.  The CatBoost model is then trained using the training set, producing a trained model which returns a single number for a given input sample. The multiclass case operates in the same fashion, but allows more than two classes to be included in the training set. 
\subsection{Likelihood Ratio Generation}\label{section:method:likelihood_ratio_generation}
After training, the entire training set is inputted to the model, producing a distribution of $N$ scores (Figure \ref{fig:method_overview}b). To be used in crafting a reliable LR system, these scores must then be calibrated so they more closely align with empirical probabilities. Machine learning outputs can suffer from both under- and over-confidence, especially when calibration is not included in the training process. A range of calibrators are available for this purpose. 
\par
We employ logistic calibration for calibrating the scores, which works by defining a sigmoidal logistic curve: 
\begin{equation}
    c(s;w,m) = \frac{1}{1 + exp(-w(s-m))},
\label{eqn:log_cal}
\end{equation}
where $s$ is the uncalibrated score produced by the model and $c$ gives the logistically calibrated score. $m$ and $w$ are parameters which are estimated from the training data to provide the line of best fit. The resulting logistic curve $c$ is thus a calibration map that transforms uncalibrated scores $s$ into calibrated scores $c(s)$. The process is visualised in Figure \ref{fig:method_overview}c, where the blue calibration curve $c$ maps the green training scores onto the diagonal. Alternative calibration methods function similarly to produce more reliable probability estimates, and the best choice of calibrator depends on the nature of your dataset. 
\par
The calibrated score $c(s)$ is also the posterior probability $p$ of $H_1$ being true for the sample. This is converted to posterior odds using the formula $p/(1-p)$. The prior odds are calculated as the ratio of the number of $H_1$-true samples and $H_2$-samples in the training set. Bayes rules is then used to produce the LR as posterior odds / prior odds \cite{leegwater_data_2024}. The values of the LRs are bounded using ELUB bounds to reduce sensitivity to extrapolation errors. ELUB bounds prevent LRs from becoming too extreme for edge values due to a lack of data points in these regions; further explanation can be found in \cite{vergeer_numerical_2016}. In the multiclass case with more than two hypotheses, we do not convert scores to LRs and instead utilise the scores directly as likelihoods for our analysis. There are additional considerations to be made when creating an LR for a mutlticlass system which can affect the precise interpretation of results, and discussion on this topic is provided in \cite{silva_filho_classifier_2023}. However, these considerations are not a focus of this work, and are left up to the practitioner.
\subsection{Likelihood Ratio Evaluation}\label{section:method:likelihood_ratio_evaluation}

For an LR system to be useful, it must be effective in two dimensions: discrimination and calibration. Discrimination is how well the system distinguishes different classes. Discrimination is typically the primary goal for a classifier, and is measured with metrics such as accuracy. Such metrics are only concerned with the top predicted class from the classifier, and do not take level of confidence into account i.e. 51\% and 100\% probabilities are equally correct. For an LR system, level of certainty is a core part of an LR's utility, hence calibration is also assessed. Calibration measures how closely the predicted LRs reflect the observed frequencies of $H_1$-true and $H_2$-true samples from the validation set, i.e. does an LR of 20 reflect a $H_1$-true sample occurring 20 times more than a $H_2$-true sample for a given point in the validation set?
\par
Best practice for evaluating LR systems is to assess discrimination and calibration in various ways \cite{banks_handbook_2020}, to verify that outputs are sensible and reasonable. Some examples of such methods are:
\begin{itemize}
    \item Pool Adjacent Violators (PAV) plot: PAV transforms the scores to optimal (in terms of $C_{llr}$) LRs using isotonic regression. The PAV plot compares the outputted LRs to these ``optimal" ones, with points ideally following the line y=x\cite{brummer_calibration_2006, vink_likelihood_2022}.
    \item Tippet plot: A Tippett plot shows the inverse cumulative densities of the log-likelihood ratio values given the two hypotheses $H_1$ and $H_2$. Ideally the LRs given $H_1$ would have all high (above 0) log-likelihood ratios, and the LRs given $H_2$ low (below 0) log-likelihood ratios \cite{ramos_validation_2020}.
    \item Empirical Cross Entropy (ECE) plot: The ECE value is a single value for a set of LRs at a particular value of the prior odds, P($H_1$)/P($H_2$). The ECE plot shows the ECE value as a function of these prior odds. The value is based on a penalty function that penalises misleading (supports wrong hypothesis) LRs. The ECE should always be lower than a completely non-informative system (outputs LR=1 for every sample). In the ECE plots included in this paper, the non-informative system is depicted as a dotted line \cite{leegwater_data_2024}.
\end{itemize}
\par
Numerically, the Log Likelihood Ratio Cost  ($C_{llr}$) \cite{brummer_application-independent_2006} captures many of the desirable components of the LR system in a single number. It is based on a strictly proper scoring rule, meaning the cost is minimised if and only if the forecast $p$ equals the true distributions of outcomes $q$. This encourages honest reporting of probabilities and cannot be ``gamed" with inaccurate likelihoods in the same way as metrics such as accuracy. The $C_{llr}$ is also application independent, since it is integrated over the priors and costs associated with different applications. It is computed as:
\begin{align}
    C_{llr} &= \frac{1}{2\Vert S_1 \Vert}\sum_{t \in S_1} \text{log}_2 ( 1 + \text{exp}(-llr_t)) \nonumber \\
    &+ \frac{1}{2\Vert S_2 \Vert}\sum_{t \in S_2} \text{log}_2 ( 1 + \text{exp}(llr_t))
    \label{eqn:cllr}
\end{align}
Where $S_1$ is the set of $H_1$-true samples, $S_2$ is the set of $H_2$-true samples, and $llr_t$ is the log-likelihood ratio under evaluation for trial $t$ (log-likelihoods are used in place of likelihoods for practical reasons). Interpretation of the value of the $C_{llr}$ is then as follows:
\begin{equation}
    C_{llr} 
    \begin{cases}
        =0:& \text{perfect} \\
        \in (0,1):& \text{useful} \\
        =1:& \text{not informative} \\
        > 1:& \text{not useful}
    \end{cases}
    \label{eqn:cllr_interpretation}
\end{equation}
Thus, $C_{llr}$ values below 1 are informative, down to a minimum of zero, which occurs for a ``perfect" LR system outputting an LR of $\infty$ for all $H_1$-true samples, and $-\infty$ for all $H_2$-true samples. $C_{llr}$ values of 1 or above are equivalent or worse than a completely non-informative system. It should be noted that beyond the above categorisation of $C_{llr}$ values, precise interpretation is difficult and must be approached carefully \cite{van_lierop_overview_2024}. 
\par
$C_{llr}$ captures loss from both the discrimination and calibration of the LR system. It is possible to separate the two components for a more detailed analysis. By recalculating $C_{llr}$ on a set of perfectly calibrated likelihood ratios an estimate can be made of the discriminative loss $C_{llr}^{min}$ of the system. The calibration loss is then the difference between the two: $C_{llr}^{cal} = C_{llr} - C_{llr}^{min}$.
\par
$C_{llr}$ can be generalised to the multiclass setting, termed Multiclass Cross-Entropy Cost ($C_{mxe}$)\cite{brummer_calibration_2006}. The formula for $C_{mxe}$ is: 
\begin{equation}
    C_{mxe} = \frac{1}{K} \sum_{k=1}^K \frac{1}{\Vert S_k \Vert} \sum_{t \in S_k} \text{log}_2 \frac{\sum_{j=1}^K \text{exp}(ll_{jt})}{\text{exp}(ll_{kt})}.
\label{eqn:cmxe}
\end{equation}
Where $K$ is the number of classes, $S_k$ is the set of samples of class $k$, and $ll_{kt}$ is the log-likelihood under evaluation for class $k$, given sample $t$ ($ll_{kt} = \text{log}_{10}P(\text{sample}\ t \vert \text{class} \ k)$). $C_{mxe}$ is equivalent to empirical cross-entropy, and degenerates to $C_{llr}$ when $K=2$. Interpretation is similar to $C_{llr}$, aside from the reference value becoming $\text{log}_2K$. For convenience, we normalise $\hat{C}_{mxe} = C_{mxe}/\text{log}_2K$ so interpretation follows that of $C_{llr}$ (eqn. \ref{eqn:cllr_interpretation}). 

% ...
\section{Results} \label{section:results}

The above described model is first evaluated in the binary setting, where the hypotheses are constructed in a one-versus-another setting. This application evaluates how well digital traces can be used in the LR system to generate LRs relating to say, the prosecution's version of events versus the defense's in a court setting, where clear cut hypotheses have already been established. The multiclass setting is evaluated next. This setting is more useful in the investigatory phase, where less is known about the sequence of events, and it is convenient to explore multiple possible events simultaneously. For this purpose, the creation of an activity timeline from digital traces is also demonstrated, to illustrate potential applications of the model. 
\par
Unless otherwise stated, subject-wise cross-validation was carried out to generate all results. From each subject, iPhone model, and carry location, multiple samples are included in the training set, with a potentially unbalanced proportion of activities. To counteract the imbalance of classes, they are weighted by their inverse frequency. The overlap in iPhone and carry location between training and validation sets also violates some of the assumptions of binomial-distribution-based analysis (according to Kruskal-Wallis tests). We therefore report results based on multilevel bootstrapping \cite{guyll_validity_2023, efron_introduction_1994, menzel_bootstrap_2017}. For a single bootstrap sample, carry locations and iPhone types are sampled with replacement. If there are $n$ unique carry locations, $n$ samples with replacement are drawn, and similarly for iPhones. All datapoints which match the iPhone -carry location combination are then used as the validation set (including resamples), and performance metrics are calculated. This is repeated 1,000 times and the mean metrics across all bootstrap samples are reported.

\begin{figure}[ht]
    \centering
    \includegraphics[width=0.45\textwidth]{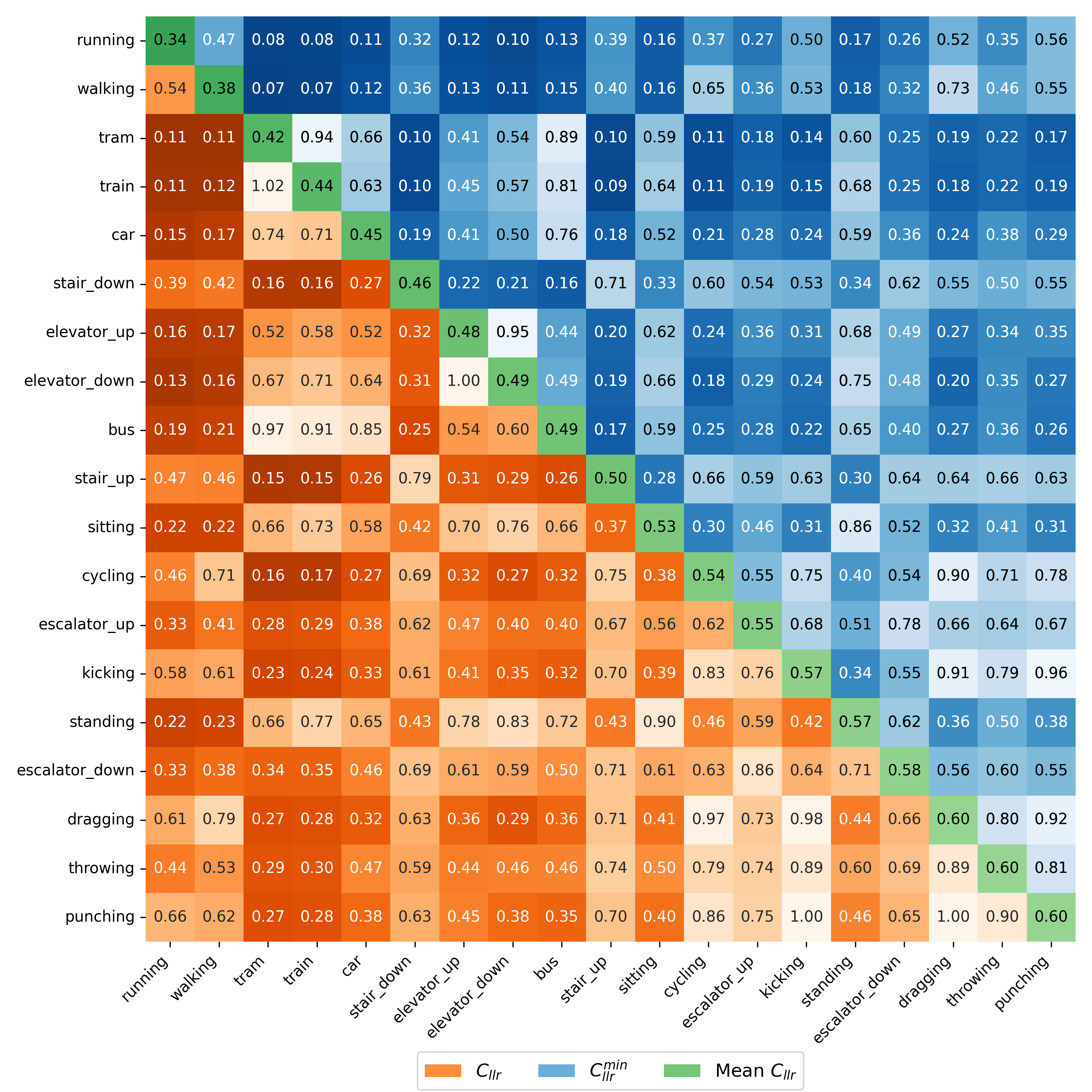}
    \caption{$C_{llr}$ and $C_{llr}^{min}$ for LR systems produced from each combination of two activity classes. Darker colours indicate lower (better) values. Each LR system is produced with $H_1$=row activity, $H_2$=column activity. Diagonal values in green are the mean $C_{llr}$ for an activity across all LR systems in which it is included.}
    \label{fig:cllr_all_acts}
\end{figure}
\begin{figure}[htbp]
    \centering
    \begin{subfigure}{0.40\columnwidth}
        \includegraphics[width=\linewidth]{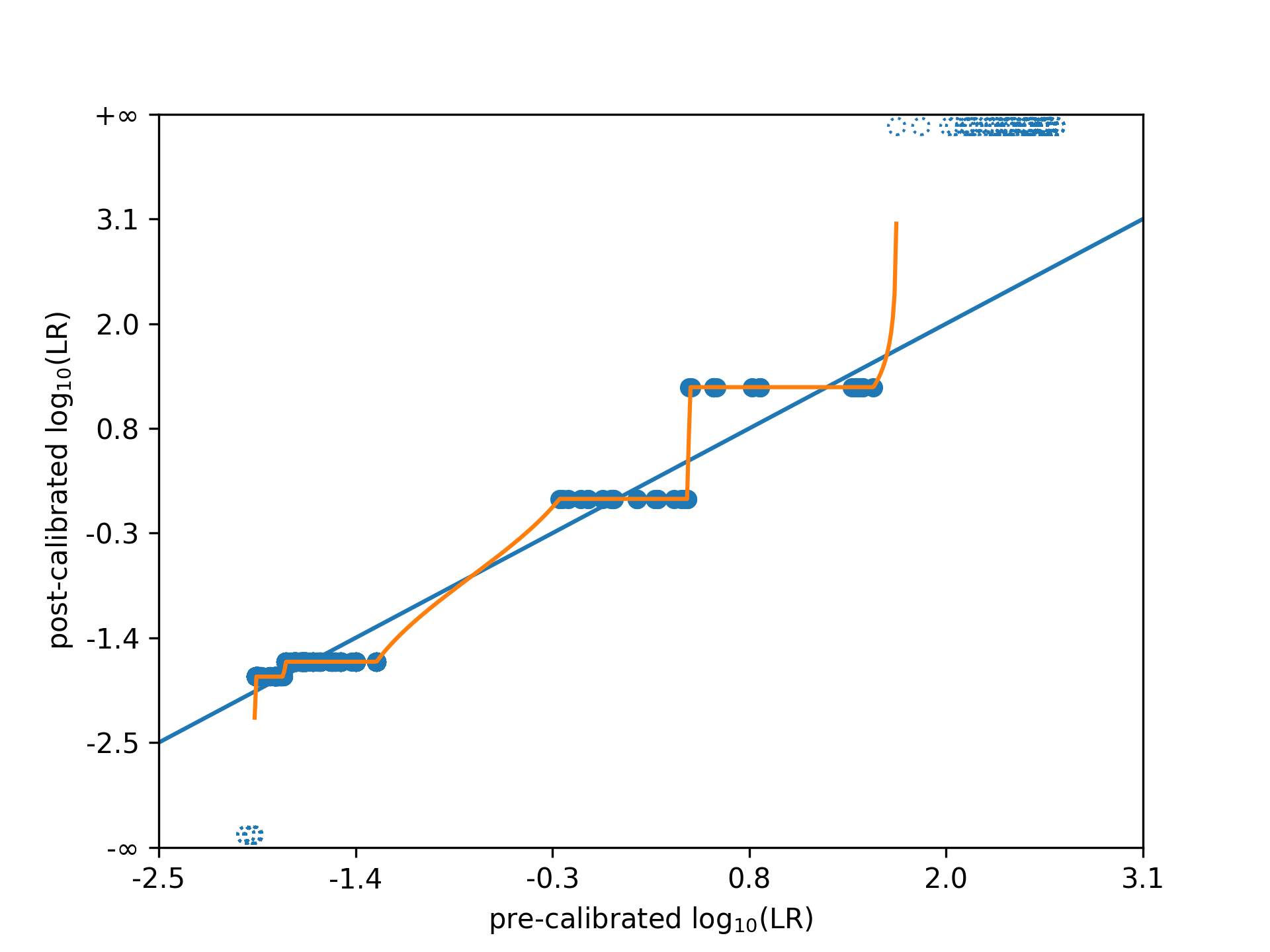}
        \caption{PAV plot (\textit{running}, \textit{tram})}
            \label{subfig:runtram_pav}
    \end{subfigure}
    \begin{subfigure}{0.40\columnwidth}
        \includegraphics[width=\linewidth]{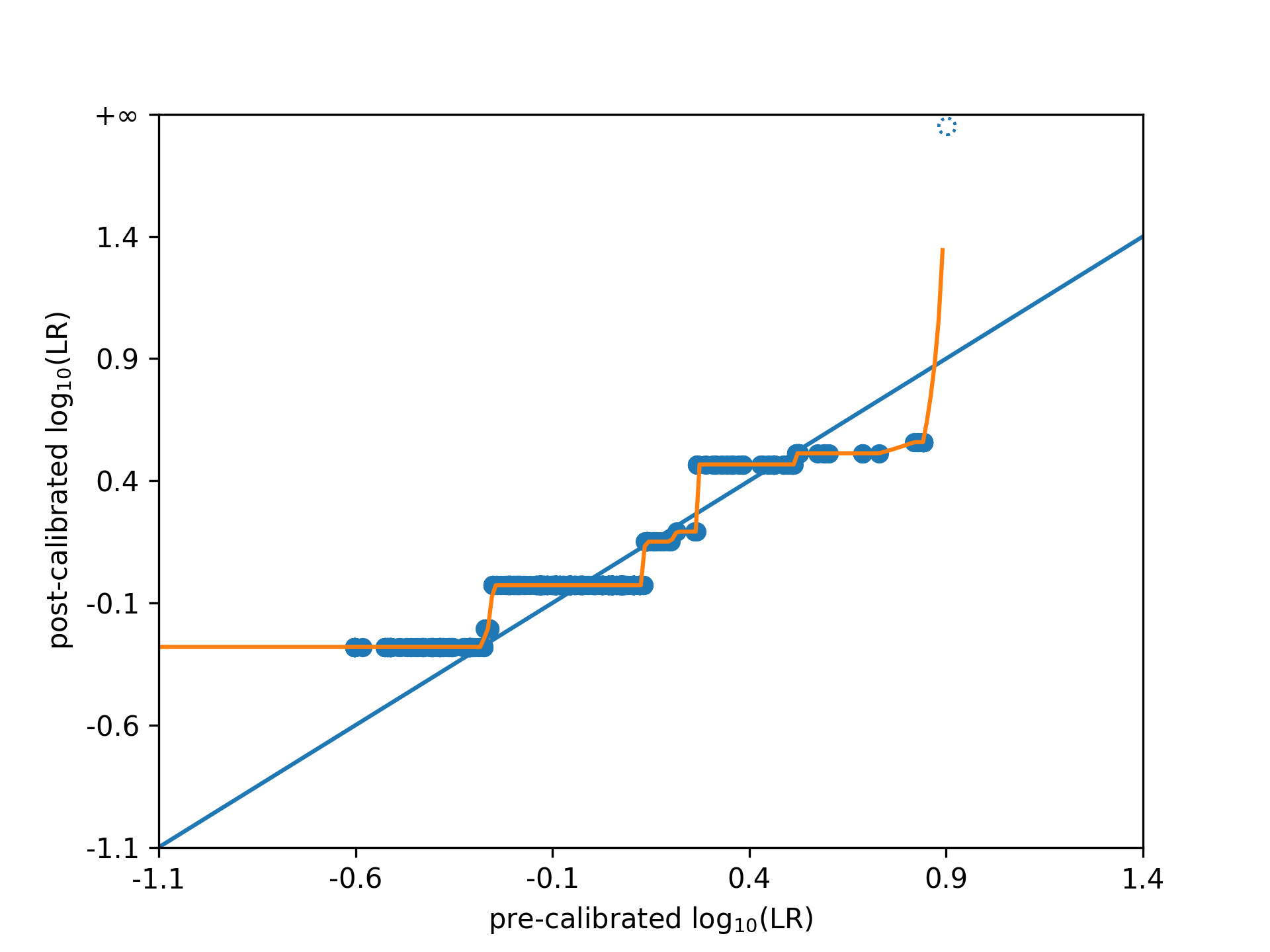}
        \caption{PAV plot (\textit{train}, \textit{tram})}
        \label{subfig:traintram_pav}
    \end{subfigure}

    \begin{subfigure}{0.40\columnwidth}
        \includegraphics[width=\linewidth]{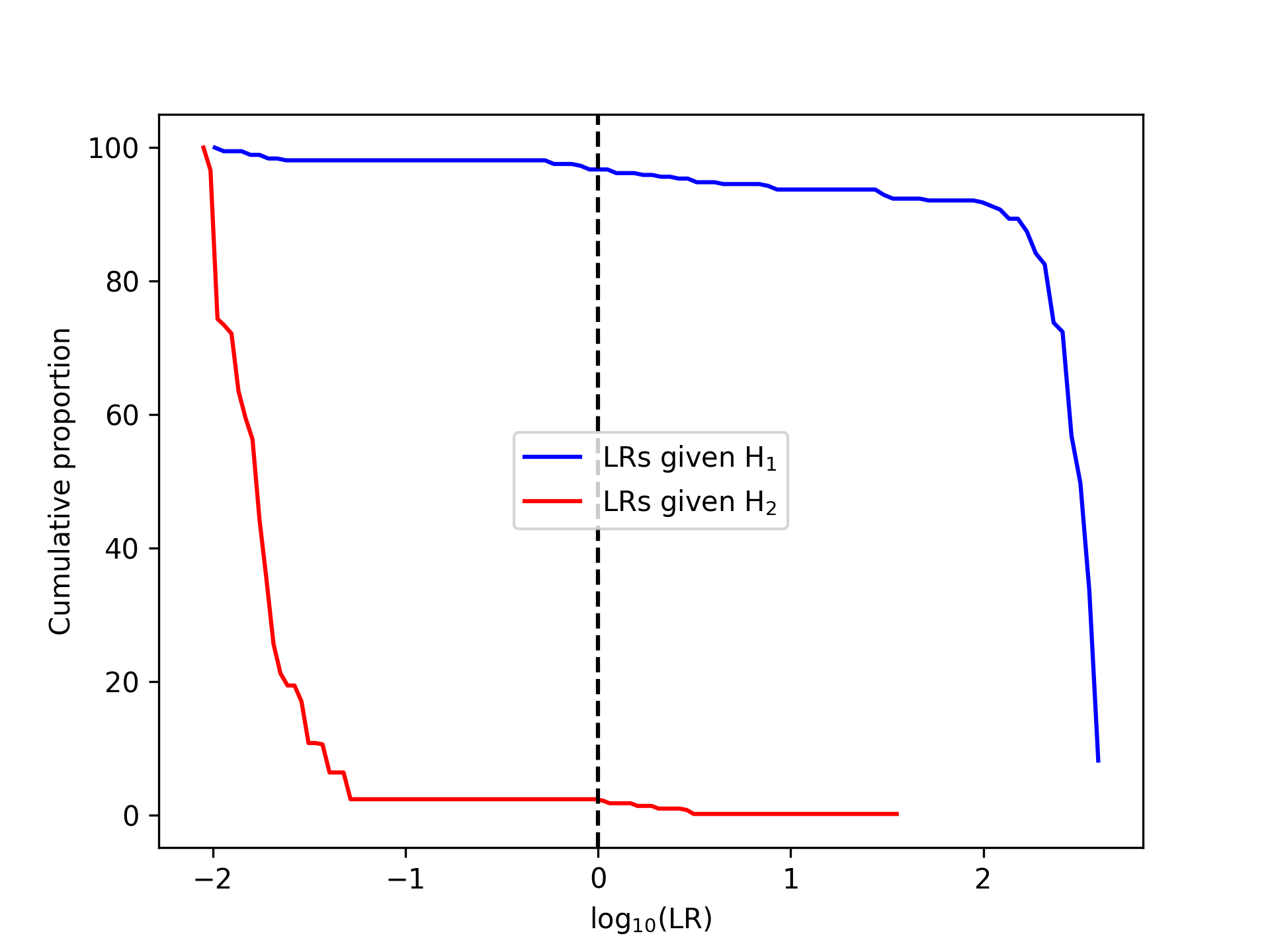}
        \caption{Tippet plot (\textit{running}, \textit{tram})}
        \label{subfig:runtram_tipp}
    \end{subfigure}
    \begin{subfigure}{0.40\columnwidth}
        \includegraphics[width=\linewidth]{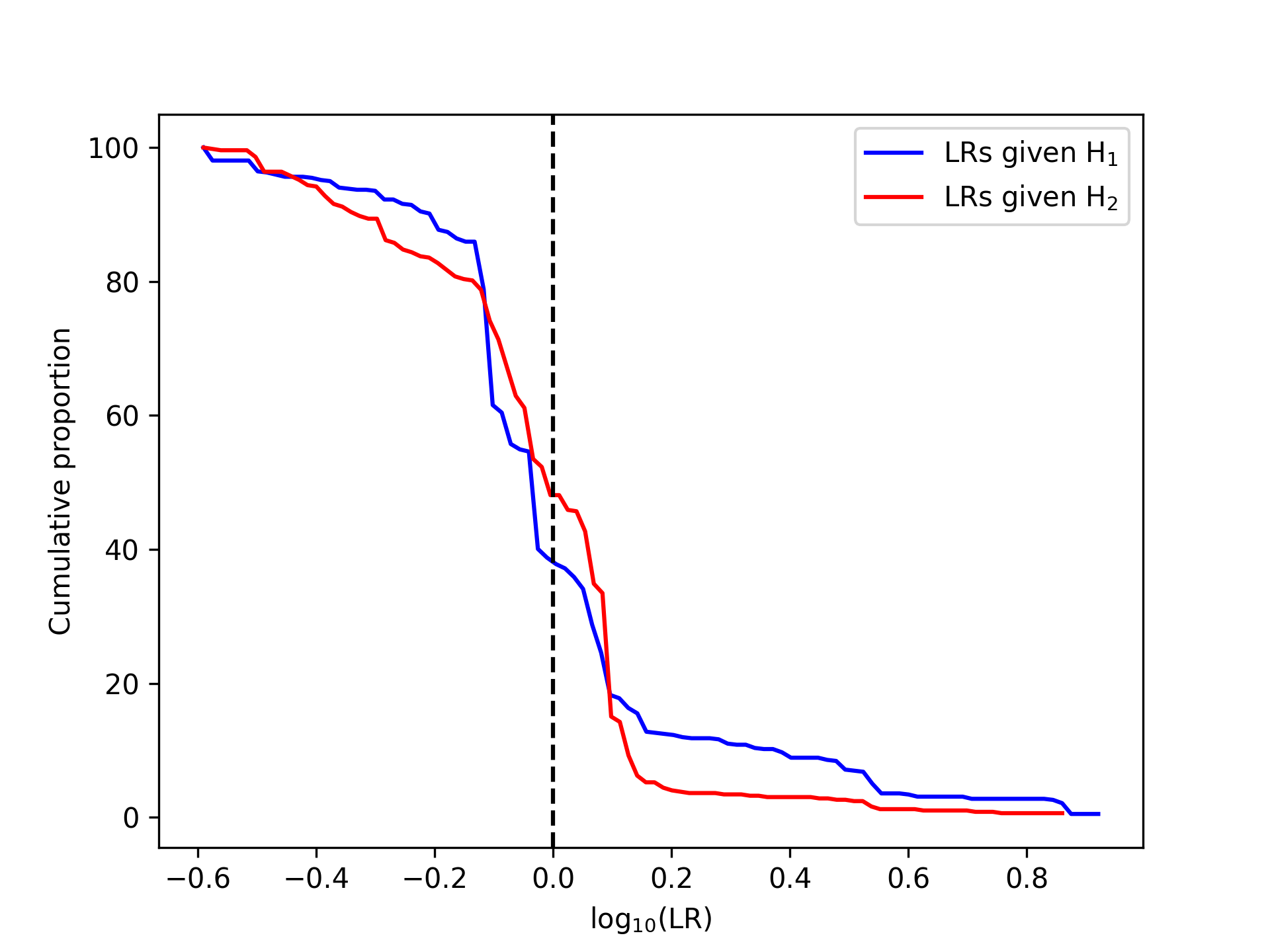}
        \caption{Tippet plot (\textit{train}, \textit{tram})}
        \label{subfig:traintram_tipp}
    \end{subfigure}

    \begin{subfigure}{0.40\columnwidth}
        \includegraphics[width=\linewidth]{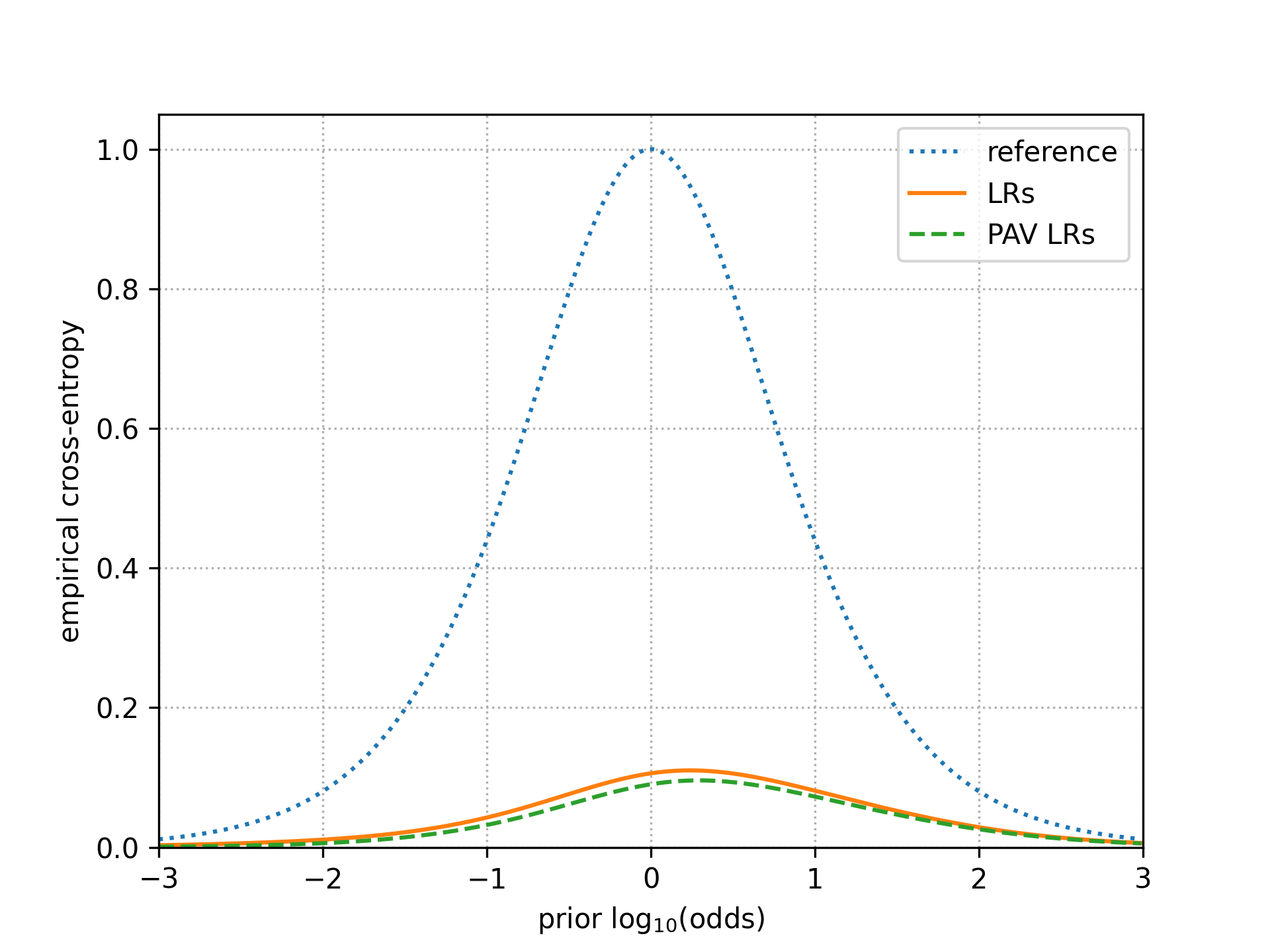}
        \caption{ECE plot (\textit{running}, \textit{tram})}
        \label{subfig:runtram_ece}
    \end{subfigure}
    \begin{subfigure}{0.40\columnwidth}
        \includegraphics[width=\linewidth]{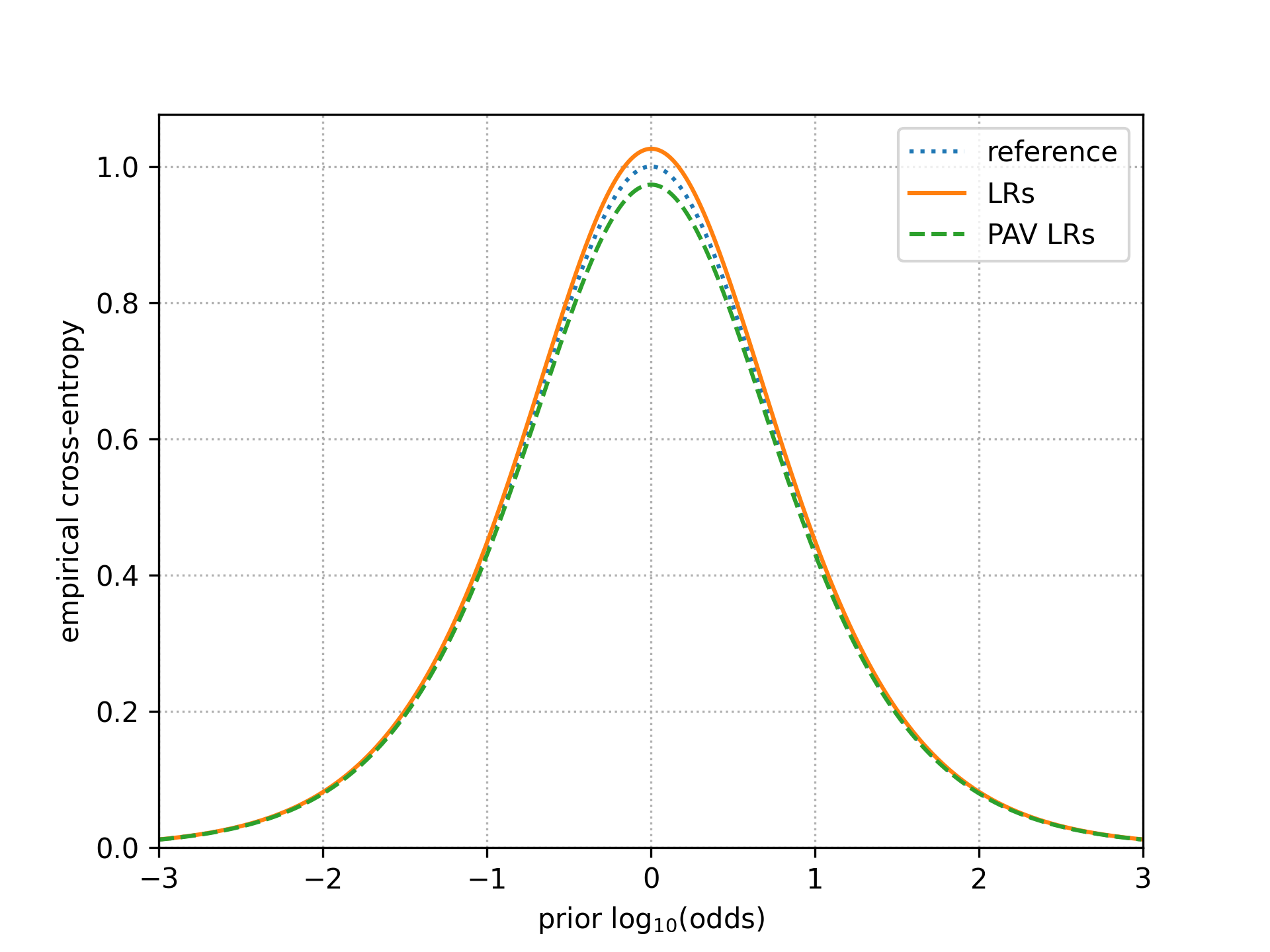}
        \caption{ECE plot (\textit{train}, \textit{tram})}
        \label{subfig:traintram_ece}
    \end{subfigure}

    \caption{Further analysis of the LR systems with the lowest (left column: (\textit{running}, \textit{tram})) and highest (right column: (\textit{train}, \textit{tram})) $C_{llr}$ values.}
    \label{fig:lrsystem_plots}
\end{figure}
\subsection{Activity Pairs \label{section:results:all_acts}}

\datasetname contains traces relating to 19 distinct activities. One of the most interesting points for a forensic investigator is to know which of these activities can be distinguished from one another using the described approach, how well they can be distinguished, and equally as important, which ones cannot be distinguished. To assess this, an LR system was created for each possible pair of hypotheses involving a single activity. The $C_{llr}$ of all systems was then computed, and results are presented in Figure \ref{fig:cllr_all_acts}. The figure consists of three sections: the lower triangle, in orange, shows the $C_{llr}$ of the system where $H_1$=activity on the row, and $H_2$=activity on the column ($H_1$ and $H_2$ are interchangeable when calculating $C_{llr}$; eqn. \ref{eqn:cllr}). The diagonal, coloured in green, shows the mean $C_{llr}$ of an activity across all eighteen LR systems it was tested in, and the upper triangle, coloured in blue, shows the $C_{llr}^{min}$ for each LR system. 
\par
For interpreting Figure \ref{fig:cllr_all_acts}, we will focus on the $C_{llr}$ values. Of all 171 possible activity pairings, 167 of the associated LR systems have a $C_{llr}$ below one, which means that these systems are at least more informative than an uninformative system, and could be of some value to an investigator. On average, \textit{running} and \textit{walking} have the lowest $C_{llr}$. We speculate that this is due to richer data availability for these activities, both from variables which measure steps directly, and others which only register when steps are detected e.g. \code{\consolas floorsAscended} \cite{van_zandwijk_have_2023}. 
\par
Highest (worst) $C_{llr}$ values occur for dynamic classes \textit{punching}, \textit{throwing}, and \textit{dragging}. This brings into question how effectively digital traces can be applied to distinguish these less typical activities. The erratic nature of the activities compared to others is likely a major source of difficulty for the systems. Although, even the worst mean $C_{llr}$ for any class (\textit{punching}) is still well within the useful range at 0.6. 
\par
The pairings with $C_{llr} > 1$ are (\textit{train}, \textit{tram}), (\textit{elevator down}, \textit{elevator up}), (\textit{punching}, \textit{kicking}), and (\textit{punching}, \textit{dragging}). The latter two pairs are pairs from the difficult to deal with dynamic category of classes (Table \ref{tab:expert_clustering}), while the former two pairs of activities have particular trouble only when paired together. This is because of the similar nature of the activities in the pairs adding confusion, discussed further in \nameref{section:results:multiclass_lrs}. 
\par
LR systems should be evaluated using a number of characteristics as well as $C_{llr}$. For practical reasons, extensive evaluation was not possible for all LR systems, but in Figure \ref{fig:lrsystem_plots}, PAV, Tippet, and ECE plots \cite{leegwater_data_2024} are presented for the best (\textit{running}, \textit{tram}) and worst (\textit{train}, \textit{tram}) LR systems according to $C_{llr}$. The left column of the figure confirms that the (\textit{running}, \textit{tram}) system has strong discriminative performance (Figure \ref{subfig:runtram_tipp}) and is well calibrated (Figures \ref{subfig:runtram_pav} and \ref{subfig:runtram_ece}). (\textit{train}, \textit{tram}) on the other hand, shows poor discrimination (Figure \ref{subfig:traintram_tipp}), but actually has close to optimal calibration (Figures \ref{subfig:traintram_pav} and \ref{subfig:traintram_ece}). This also agrees with the values in Figure \ref{fig:cllr_all_acts}; $C_{llr}=1.02$ is primarily from the discrimination loss $C_{llr}^{min}=0.94$, with a calibration loss of only $C_{llr}^{cal}=0.08$. 
\par
Figure \ref{fig:cllr_all_acts} can be used as a reference in the early stages of analysis, to help assess whether digital traces could be helpful in a specific case. 
\subsection{Ablation \label{section:results:ablation}}
\begin{table}
\centering
\scalebox{0.65}{
\begin{tabular}{c|c|c|c|c|c|c|c}
\hline
\multirow{2}{*}{Scorer}  & \multirow{2}{*}{Calibrator} & \multirow{2}{*}{Accuracy $\uparrow$} & \multirow{2}{*}{Mean $C_{llr} \downarrow$}
    & \multicolumn{4}{c}{\% $C_{llr}s$ below $\uparrow$}  \\ 
    \cline{5-8}
    & & & & 1.00  & 0.75 & 0.50 & 0.25 \\
\hline
\hline
\multirow{3}{*}{CatBoost}
    & LogReg & 85.0 & 0.504 & 97.7 & 84.8 & 52.6 & 14.6  \\
    & KDE & 85.0 & 0.511 & 97.1 & 82.5 & 52.6 & 14.0  \\
    & Gauss & 85.0 & 0.546 & 97.1 & 80.1 & 47.4 & 10.5  \\
\hline
\multirow{3}{*}{Random Forest}
    & LogReg & 86.6 & 0.609 & 85.4 & 66.1 & 47.4 & 18.1  \\
    & KDE & 86.7 & 0.617 & 84.2 & 71.3 & 53.2 & 18.1  \\
    & Gauss & 86.7 & 0.612 & 91.8 & 74.9 & 40.9 & 10.5  \\
\hline
\multirow{3}{*}{XGBoost}
    & LogReg & 86.2 & 0.760 & 68.4 & 55.6 & 39.2 & 14.6  \\
    & KDE & 86.2 & 0.871 & 64.3 & 52.6 & 31.6 & 9.9  \\
    & Gauss & 86.2 & 0.787 & 67.3 & 55.6 & 34.5 & 8.2  \\
\hline
\multirow{3}{*}{Decision Tree}
    & LogReg & 83.3 & 1.305 & 42.7 & 24.6 & 8.2 & 0.6 \\
    & KDE & 83.3 & 1.303 & 41.5 & 25.1 & 8.2 & 0.6  \\
    & Gauss & 83.3 & 1.261 & 45.0 & 28.1 & 9.9 & 0.6  \\
\hline
\end{tabular}
}
\caption{Ablation study results on all pairs of activities. LogReg=Logistic Regression, KDE=Kernel Density Estimation, Gauss=Gaussian calibrator. }
\label{tab:ablation_results}
\end{table}

The described LR system consists of two components: a scorer and a calibrator. For each component, multiple options are available. In Table \ref{tab:ablation_results}, we compare the performance of twelve methods of producing LR systems by creating an LR system for each possible activity pairing and calculating the accuracy and $C_{llr}$. The mean values are presented, as well as the proportion of activity pairings with $C_{llr}$ below the thresholds 1.00, 0.75, 0.50, and 0.25. 
\par
The combination of a CatBoost scorer and Logistic Regression calibrator performs the best. The mean $C_{llr}$ of all systems is 0.504, and 97.7\% of all systems are more informative than the reference, with over half of all systems having a $C_{llr}$ below 0.5. The only category in which this combination is not best overall is in percentage of $C_{llr}$s below 0.25, where Random Forest combined with Logistic Regression or Kernel Density Estimation is better (14.6\% versus 18.1\%). While this does indicate that Random Forest may have a higher ceiling for quality of LR systems, in the majority of cases CatBoost + Logistic Regression performs best.
\par
Accuracy measures the percentage of correct classifications made by the scorer, which is the typical goal for a machine learning classifier. Interestingly, both Random Forest and XGBoost are better classifiers than  CatBoost, but have worse $C_{llr}$. This demonstrates that when choosing a scorer, it is important to not just select the best classifier, since its performance in an LR system may be inferior to a weaker classifier with better calibration.
\begin{figure}[ht]
    \centering
    \includegraphics[width=0.48\textwidth]{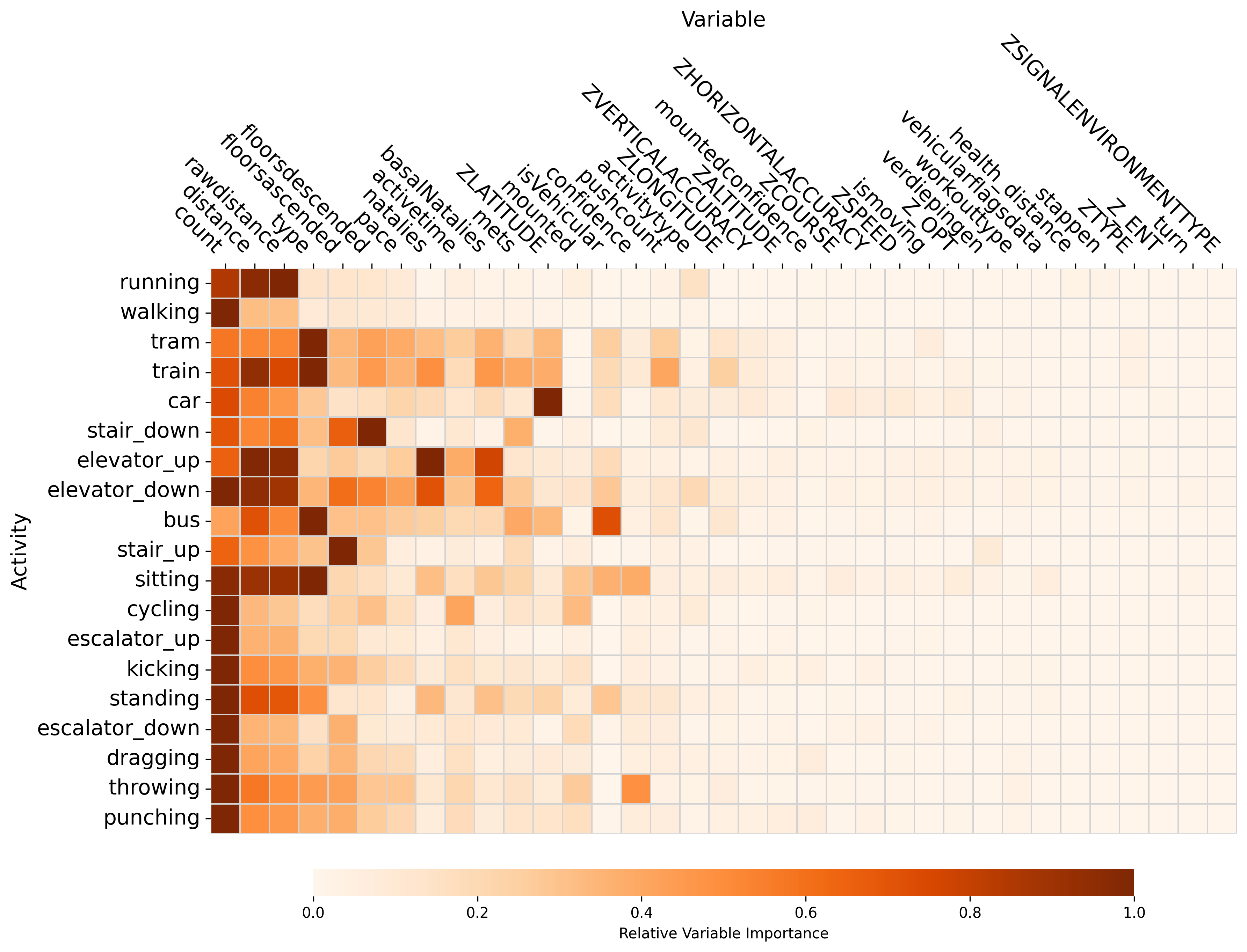}
    \caption{Heatmap displaying average variable importances for each activity class.}
    \label{fig:feat_imp}
\end{figure}
\subsection{Variable Importance \label{section:results:feat_imp}}

An advantage of using Catboost is that it allows us to inspect the importance of each input variable when classifying, and how they vary between classes. Importance was calculated using PredictionValuesChange importance \cite{prokhorenkova_catboost_2018}, which measures how predictions shift when variables are perturbed. The importance values for all eighteen pairings involving a particular class were aggregated and normalised, and the results are displayed in Figure \ref{fig:feat_imp}. In this figure, we can inspect which variables contribute most to the classification of each activity class. Variables are arranged in descending order of average importance, and importance values are normalised row-wise. This analysis was also repeated weighting importance by the inverse $C_{llr}$ of the pairing, to reduce contribution of non-informative classifiers, however this did not have any material impact on results. 
\par
Variable \code{\consolas count} is one of the top most important variables for most of the activities in \datasetname, which has been identified in previous work as the cumulative number of steps \cite{van_zandwijk_phone_2021}. Alongside variables \code{\consolas distance} and \code{\consolas rawdistance}, these variables have been previously studied for forensic applications \cite{van_zandwijk_phone_2021}. \code{\consolas count} in particular is quite informative for activity classification, and merits further targeted research. Many variables contribute little to classifying any of the activities. 17 out of the 35 variables average less than 3\% of the importance of the largest recorded value, most of which can be attributed to noise. These variables can be removed to reduce model size without a significant loss of prediction quality.  
\par
Distributions of variable importance is interesting to observe, with marked difference between activities. For example, \textit{running} has variables \code{\consolas count}, \code{\consolas distance}, and \code{\consolas rawdistance} as its most important, none of which are the most important for \textit{bus}, whose top variable is \code{\consolas type}, which has previously been identified as being related to travelling in a vehicle \cite{van_zandwijk_phone_2021}. Such interactions between variables and activities can be instructive for future research further connecting these variables with real-world quantities. 
\subsection{Sensitivity Analysis \label{section:results:sens_analysis}}


Above results are calculated using training and test sets without any overlap in subjects, to better assess generalisability to new subjects. However, there are two additional factors which can also affect generalisability: phone type and carry location. To investigate how results change when analysis is performed on a new phone/carry location, the evaluation procedure was repeated, with the alteration that after the training/validation set had been constructed for one subject fold of an activity pair, all samples from one of the factors (e.g. carry location: hand), were removed from the training set, and all samples \textbf{not} from that factor were removed from the test set. This produced a training and test set with no overlap in both subjects and that factor. This was repeated for every factor on each subject fold-activity pair combination, with the same LR system generation and evaluation being performed. The change in performance compared to training/test sets only keeping subjects distinct was then scrutinised. To remove the effect coming from reduced sample size, the procedure was also repeated with training/test set size reduced by removing samples at random to match the size of the modified sets, but without separation of the factors.
\par
For carry location, a one-sided Wilcoxon signed-rank test showed no significant increase (p-value=1.00) in $C_{llr}$ between separating by carry location and not. Performance was consistent when the tested data came from a carry location not present in the training set.
\par
For phone type, the Wilcoxon signed-rank test showed a significant increase in $C_{llr}$ with a p-value $\ll 0.01$. $C_{llr}$ increased by a mean value of 0.04 when testing on a new phone type. Overall, 56\% of $C_{llr}$s increased by 0.05 or less, 84\% by 0.10 or less, and 96\% by 0.15 or less. The largest single increase in $C_{llr}$ was for the pairing (\textit{car}, \textit{elevator up}), increasing 0.19 from 0.50 to 0.69. Across all pairings, only two crossed the $C_{llr}=1$ threshold to an uninformative system as a result of separating phone type, namely (\textit{dragging}, \textit{cycling}) and (\textit{dragging}, \textit{kicking}), both of which already had $C_{llr} > 0.95$. 
\par
Overall, it can be stated that performance is expected to worsen if the data being tested is from a phone not present in the training set. However, no such consideration must be taken in regards to carry location.  
\begin{figure}[ht]
    \centering
    \includegraphics[width=0.45\textwidth]{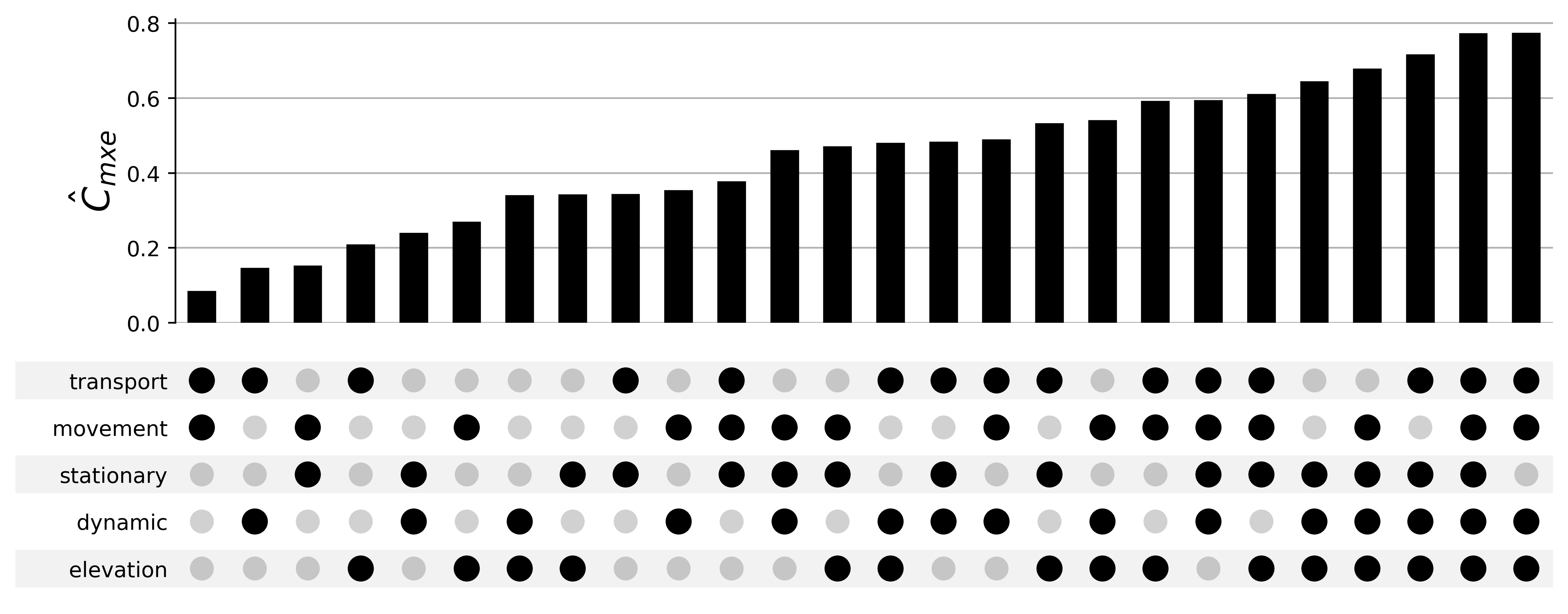}
    \caption{UpSet plot\cite{lex_upset_2014} of $\hat{C}_{mxe}$ for all unique combinations of activity groupings. }
    \label{fig:upset}
\end{figure}

\begin{table}
    \centering
    \scalebox{0.85}{
    \begin{tabular}{c|c}
        \hline
       Group & Activity Types\\
       \hline
       Movement  & cycle, run, walk \\
       Transport  & bus, car, train, tram \\
       Dynamic  & drag, kick, punch, throw \\
       Elevation  & elevator, escalator, stair\\
       Stationary  & sit, stand \\
       \hline
    \end{tabular}
    }
    \caption{Semantic grouping of activity classes. Elevation activities include both up and down variations.}
    \label{tab:expert_clustering}
\end{table}

\subsection{Multiclass LR systems \label{section:results:multiclass_lrs}}

The proposed approach can be theoretically extended to as many hypotheses as there are classes in the training set. This may be of interest to an investigator in the earlier stages of an investigation, when not enough information is yet known to reduce the set of possible activities to only two options, and they wish to explore the likelihoods of multiple hypotheses simultaneously. 
\par
The na{\"i}ve approach to a multiclass system is to simply include all activities in the training set and run the resulting system on your data. However, as seen in Figure \ref{fig:cllr_all_acts}, due to the overlapping nature of certain activities, some classes are mutually confused due to high similarities e.g. elevator up and elevator down. Including both of these activities as output options can cause the estimated likelihood of both to be suppressed, affecting performance. When na{\"i}vely using all nineteen of the \datasetname activities in a single LR system, the resulting $\hat{C}_{mxe}$ is 0.96, close to uninformative. It therefore makes sense to group similar activities into distinct clusters, to avoid misleading deflation of likelihood values. This has the added benefit of forcing fewer restrictions on the investigator when selecting what activities are of interest. 
\par
It was found that grouping classes into the five categories which were used when creating the dataset performed as well or better than any tested unsupervised clustering method (see Figure \ref{fig:clustering_methods} in the Appendix). Using these groupings also provides semantic meaning, aiding interpretation. The groups are listed in Table \ref{tab:expert_clustering}. In this set up, an investigator may be interested in knowing whether a user was taking transportation, moving themselves, or stationary during a period of time. They could then select these three groups in the multiclass system and inspect the resulting likelihoods.
\par
Similarly to the binary case, some sets of groupings will be easier to distinguish than others. In Figure \ref{fig:upset}, the $\hat{C}_{mxe}$ of the LR systems of all combinations of the activity groupings is displayed. Systems containing only two groups perform best, and $\hat{C}_{mxe}$ increases with number of included groups. All systems are better than the reference, with the two worst (all groups included and all but stationary included) having a $\hat{C}_{mxe}$ of 0.78. 
\begin{figure}[ht]
    \centering
    \includegraphics[width=0.45\textwidth]{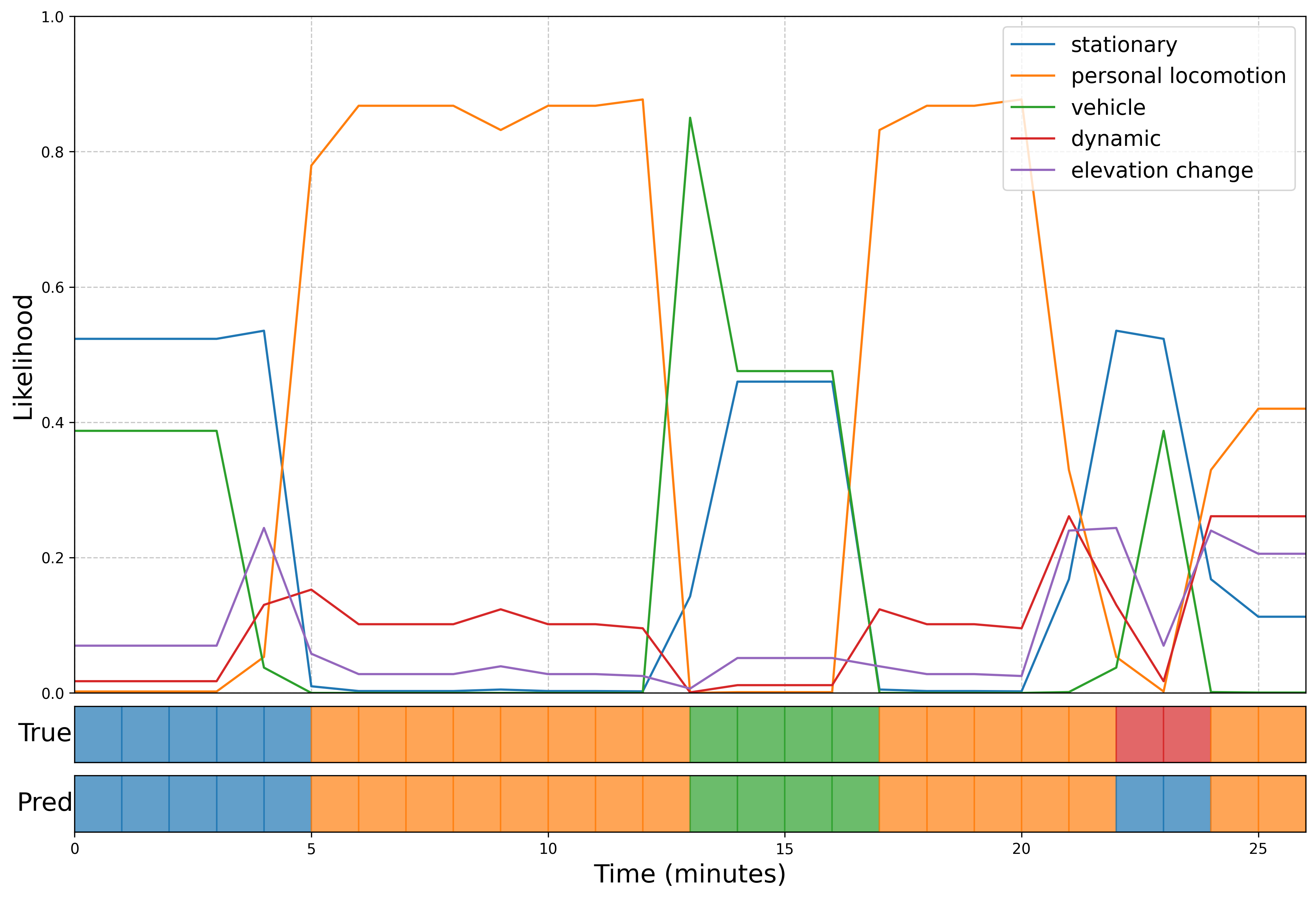}
    \caption{Example timeline illustrating change in likelihood for each activity grouping as ground truth changes over time. Ground truth class and highest likelihood predicted class at each timestamp displayed at the bottom.}
    \label{fig:timeline}
\end{figure}
\subsection{Timelines} \label{section:results:timelines}

Another application of the model is the creation of activity timelines from digital trace data. Establishing a timeline can be an instructive part of an investigation, and using our approach, one can be directly created in order to inspect the likelihood of different activities/activity groupings over time. Figure \ref{fig:timeline} shows a timeline for a 26 minute period in which a person was sitting, walked to the tram, walked for a few minutes after getting off the tram, kicked/punched someone for 2 minutes, then ran away, constructed by reordering samples from \datasetname. The timeline was constructed using the \datasetname dataset, extracting and reordering the datapoints from the validation set to follow the described pattern. The bottom row of squares shows the most likely class according to the model, with the correct group in the row above. It can be seen that the timeline is mostly accurate, correct for 24 of the 26 minutes, all correct for the groups \textit{stationary}, \textit{personal locomotion}, and \textit{vehicle}. The \textit{dynamic} group is wrongly identified and assigned a low likelihood. As in the binary case, identifying this group of activities is challenging. 
\section{Discussion} \label{section:discussion}



The proposed approach to translate digital traces to LRs was tested and evaluated using iPhone traces from the \datasetname dataset. iPhones were studied in this work because they have a consistent file storage system compared to the diverse ecosystem of Android OS's and are the most extensively studied, in particular in relation to physical activities. As more traces become available they can be inserted directly into our approach without adjustment. Currently, the results only pertain to digital traces from iPhones and their usefulness in an LR system.
\par
The system was first analysed in the binary case, comparing the likelihood of the traces being generated by either one of two activities. It was found that according to the $C_{llr}$ values, 167 pairs of activities could be used to produce an LR system which is more informative than a baseline reference system. Additional analysis was carried out on the best and worst systems, (\textit{running}, \textit{tram}), and (\textit{train}, \textit{tram}), respectively, which agreed with the initial conclusions from the $C_{llr}$ and $C_{llr}^{min}$ values. For use in actual casework, however, a fuller analysis would need to be carried out on any generated LR system, and its quality would have to be decided on a case-by-case basis. Such deep analysis was impractical to include given the volume of pairings, and the primary concern was assessing in which scenarios digital traces could be helpful to an investigator. The heatmap (Figure \ref{fig:cllr_all_acts}) is suggested to be used as an initial reference, to help inform whether it is worth looking into creating an LR system from digital traces, and expert judgement and case specific information should always be the deciding factors. 
\par
Investigations into variable importance using our machine learning approach can benefit future research into the meaning and application of digital traces. Additionally, the importance could lead researchers towards creating some rule-based systems for activity classification, which may be preferable in certain situations. For example, variable \code{\consolas type} is of high importance for classes \textit{tram}, \textit{train}, and \textit{bus}, supporting previous findings linking the variable with transport activities. Finding a reliable link between this variable and these activities could prove fruitful.
\par
An important point of consideration for using the proposed approach in investigations is the generalisability of results to the situation under study. It was shown that carry location does not need to be a concern when applying the approach, which is helpful since this information may be hard to discern in an actual case. However, the sensitivity analysis did reveal that phone type mismatches between the training and test set can adversely impact the quality of results. If the digital traces are taken from a phone not present in \datasetname, then the LR system trained on \datasetname is expected to have a $C_{llr}$ 0.04 higher than what is presented here. \datasetname contains data from four iPhones that were relevant at the time of collection. As new phones and operating systems are continually being released, the dataset will require regular updating to provide maximum benefit and performance. Such continual updating is a considerable task, and therefore cooperation between agencies is required to keep valuable forensics datasets up to date.  The outlined approach can also work on an entirely new dataset of digital traces without overlap with \datasetname, relevant to the case at hand. If it is not possible to include the targeted phone and operating system in the training set, the investigator should take care when evaluating their LR system and be aware of the associated reduction in effectiveness. 
\par
The model was also extended to multiclass scenarios, where the likelihood that the traces were generated while one of multiple activities was being performed could be estimated. Multiclass is obviously a generalisation of the binary case; however, due to the drop in precision, it is recommended for use in the investigatory phase, and in practice, a binary setup would be presented for evidence. Multiclass is most useful to explore many possibilities at once in the early stages of an investigation, and further refinement is up to the expert. It was also found to be necessary to group activities into meaningful clusters, both to improve results and make analysis easier. The expert can begin by using all groups if desired, but as demonstrated in Figure \ref{fig:upset}, reducing the number of investigated groups improves the utility of the system. Reduction of groups could be in the form of outside knowledge, for example, if the period in question is too late for public transport, group \textit{transport} could be ruled out, or if \textit{dynamic} activities are not relevant to the case, they can also be omitted. It should be noted that multiclass results are presented using likelihoods, rather than LRs. The likelihoods can be transformed to LRs quite easily, such as using a one-versus-all approach, but in our analysis we found using likelihoods more instructive. The creation of activity timelines from digital trace data was also demonstrated as one such application of multiclass activity scenarios. 
\section{Conclusions and Future Work} \label{section:conclusion}

In this work we explored what benefit digital traces from smartphones could bring to forensic activity classification. Our outlined method provides a straightforward way to translate digital traces into LRs for activities, both in binary and multiclass settings. This is the first published use of these digital traces to classify activities, rather than validating the accuracy of the digital traces' proprietary labelling, and it was found that 167 possible pairs of activities could be distinguished. This analysis was facilitated by the collection of dataset \datasetname, containing a large and diverse set of digital traces labelled with nineteen distinct activities, which has also been made public. 
\par
The most important next step for improving how forensic practitioners can use digital activity traces is data collection. \datasetname is the first public dataset containing forensically relevant digital traces from iPhones, representing a large amount of experimental data and activities. However, as new models continue to be released, it is important for the dataset to be updated to maintain effectiveness. Additionally, data collected from new subjects and locations will further enhance the robustness of models trained on that data. 
\section{Acknowledgements}
We would like to thank Loes Quirijnen for planning and collecting all the data for \datasetname. We also thank Marjan Sjerps for offering her time during discussions and providing valuable feedback on this paper, alongside Wauter Bosma and Pim Meulensteen.

\bibliography{references}
\appendix
\section{Appendix} \label{section:appendix}
\renewcommand{\thefigure}{\arabic{figure}}
\begin{figure}[ht]
    \centering
    \includegraphics[width=0.4\textwidth]{figs/images/clustering_methods_comparison.png}
    \caption{$\hat{C}_{mxe}$ values for different clustering methods as number of clusters is varied. Knowledge-based approaches \textbf{expert} and \textbf{chat-gpt} are in green.}
    \label{fig:clustering_methods}
\end{figure}

\end{document}